\documentclass[runningheads]{llncs}
\usepackage{eccv}
\usepackage{eccvabbrv}
\usepackage{array}
\usepackage{graphicx}
\usepackage{booktabs}
\usepackage{multirow}
\usepackage{makecell}
\usepackage{tabularx}
\usepackage{enumitem}
\usepackage{fvextra}
\usepackage{placeins}

\newcolumntype{C}[1]{>{\centering\arraybackslash}m{#1}}
\newcolumntype{Y}{>{\centering\arraybackslash}X} 
\newcolumntype{Z}{>{\raggedright\arraybackslash}X} 

\usepackage[accsupp]{axessibility}  
\usepackage{hyperref}
\usepackage{orcidlink}

\begin{document}
\title{Omni-RRM: Advancing Omni Reward Modeling via Automatic Rubric-Grounded Preference Synthesis}
\titlerunning{Omni-RRM}

\author{
Zicheng Kong\inst{1}\textsuperscript{*} \and
Dehua Ma\inst{1}\textsuperscript{*} \and
Zhenbo Xu\inst{1,4} \and
Alven Yang\inst{1} \and
Yiwei Ru\inst{1} \and
Haoran Wang\inst{2} \and
Zixuan Zhou\inst{1} \and
Fuqing Bie\inst{1} \and
Liuyu Xiang\inst{1} \and
Huijia Wu\inst{1} \and
Jian Zhao\inst{3} \and
Zhaofeng He\inst{1}\textsuperscript{\dag}
}

\authorrunning{Z.~Kong et al.}

\institute{
Beijing University of Posts and Telecommunications, Beijing, P.R. China \and
Tsinghua University, Beijing, P.R. China \and
Institute of Artificial Intelligence (TeleAI), China Telecom, P.R. China \and
ShiFang Technology Inc., Hangzhou, P.R. China\\[0.4em]
\email{zhaofenghe@bupt.edu.cn}\\
\textsuperscript{*}Equal contribution. \textsuperscript{\dag}Corresponding author.
}

\maketitle

\begin{abstract}
Multimodal large language models (MLLMs) struggle with alignment due to the limitations of existing reward models (RMs), which are predominantly vision-centric, dependent on costly human labels, and provide opaque scalar scores that fail to capture nuanced reasoning, leading to brittle alignment. We present Omni-RRM, an \textbf{Omni}-modal \textbf{R}ubric-grounded  \textbf{R}eward  \textbf{M}odel that generates multi-dimensional reward signals across text, image, video, and audio. To overcome the high cost and inherent inconsistency of human-centric evaluation in multi-dimensional reasoning, we introduce \textbf{Omni-Preference}, a high-quality dataset constructed via automatic rubric-grounded preference synthesis. In this pipeline, teacher models reconcile raw preferences into explicit justifications, ensuring that the synthesized supervision is both high-fidelity and interpretable. Omni-RRM is trained using a progressive SFT + GRPO regimen, specifically optimized to sharpen reward discrimination on low-margin, hard preference pairs. It achieves state-of-the-art accuracy on video (80.2\% on ShareGPT-Video) and audio benchmarks (66.8\% on Audio-HH-RLHF and 65.0\% on TA2T), yielding a five-benchmark Overall accuracy of 70.4\% and a +17.0\% relative gain over its backbone. Furthermore, Omni-RRM effectively guides Best-of-$N$ selection and exhibits robust transfer to text-only alignment. All resources, including the dataset, training and inference code, and model checkpoints are available at \url{https://tmfk418.github.io/Omni-RRM}.
\keywords{Multimodal alignment \and Reward modeling \and Rubric-grounded preference}
\end{abstract}

\section{Introduction}
\label{sec:intro}

Multimodal large language models (MLLMs) such as GPT-4o, Gemini 1.5, and Qwen2.5-Omni are rapidly advancing in their ability to process and reason over complex multimodal inputs \cite{openai2024gpt4o,gemini2024,xu2025qwen25omni}. Yet, achieving consistent reliability and high-fidelity reasoning remains a significant challenge, since raw capability improvements do not automatically yield trustworthy outputs across modalities. This is frequently manifested in issues such as brittle visual grounding, temporal hallucinations in videos, or unfaithful audio interpretations \cite{liu2023visualinstructiontuning,chen2023internvl,bai2023qwenvl}. A central component in addressing these gaps is the reward model (RM): it serves as a provider of critical discriminative reward signals, offering training-time supervision for post-training (e.g., preference optimization) and inference-time guidance for performance enhancement (e.g., reranking). From a principal--agent perspective, reward models act as incentive functions through which researchers encode task-specific requirements and nuanced qualitative preferences; consequently, the information density and structural clarity of these reward signals largely determine how effectively a model's potential can be steered, audited, and scaled.

Despite their importance, current multimodal RMs remain limited in both signal granularity and modality coverage. Most existing models are still largely vision-centric and compress nuanced preferences into single scalar scores, resulting in sparse reward signals that offer little criterion-level transparency for debugging or controllable improvement \cite{kirstain2023pickapic}. While recent reasoning reward models begin to provide free-form textual critiques or chain-of-thought rationales \cite{xiong2025llava,wang2025unifiedrewardthink,jin2025omnireward}, such reward signals are typically not grounded in an explicit, fixed rubric, making dimension-wise assessments harder to audit and less consistent across modalities. Moreover, high-quality preference supervision still heavily relies on expensive human annotation, which creates a bottleneck for rapidly adapting reward modeling to under-explored modalities such as audio \cite{kaufmann2023rlhf,sun2024aligning}. These gaps motivate a scalable and efficient reward modeling framework that: (i) can construct high-fidelity preference supervision in an automated manner, and (ii) provides rubric-grounded reward signals that are unified yet modality-aware.

To overcome these barriers, we introduce \textbf{Omni-RRM}, an open-source, \emph{rubric-grounded} reward model. The term \textbf{omni} highlights its versatility across \textbf{text, image, video, and audio}.
Distinct from opaque scalar RMs, Omni-RRM enforces the generation of structured, rubric-grounded justifications prior to the final verdict. This design introduces structural regularization that encourages the reward model to align its preferences with human-auditable criteria, shifting reward modeling from a `black-box' scoring task toward a more auditable criterion-level judgment process.

To achieve this, we first propose an automated preference synthesis pipeline that generates high-fidelity supervision without human annotation. Using this pipeline, we construct Omni-Preference, a large-scale multimodal dataset where candidate pairs are generated via capability-contrast sampling and annotated with teacher-reconciled rationales. Omni-RRM is then trained through a progressive SFT-RL regimen: initial supervised fine-tuning internalizes the rubric schema, followed by Group Relative Policy Optimization (GRPO) to sharpen discrimination on nuanced, low-margin pairs. Empirically, Omni-RRM achieves state-of-the-art preference accuracy on video (80.2\%) and audio benchmarks (66.8\% on Audio-HH and 65.0\% on TA2T), reaching 70.4\% five-benchmark Overall accuracy with a +17.0\% relative gain over its backbone. It also effectively guides Best-of-$N$ selection and exhibits positive transfer to text-only benchmarks, proving its versatility as a unified reward source.

In summary, our contributions are three-fold:
\begin{itemize}
    \item \textbf{Automatic Rubric-Grounded Preference Synthesis.} We propose a fully automated pipeline for preference synthesis that overcomes the scalability and consistency bottlenecks inherent in manual multimodal annotation. By leveraging teacher-reconciled consensus, this pipeline systematically generates \textbf{Omni-Preference}, a diverse omni-modal dataset enriched with high-fidelity, rubric-grounded rationales. This approach encourages cross-modal reward calibration and provides dense, discriminative supervision that facilitates more robust preference modeling across modalities.
    \item \textbf{Omni-RRM.} We develop \textbf{Omni-RRM}, an omni-modal reward model supporting text, image, video, and audio. By enforcing a shared rubric within a single omni-modal framework, Omni-RRM encourages cross-modal reward calibration. It is optimized via a progressive SFT + GRPO regimen, utilizing a rubric-grounded reward design to generate interpretable, rubric-grounded reward signals that are consistent and auditable across modalities.
    \item \textbf{Extensive Validation.} Omni-RRM-7B achieves state-of-the-art accuracy among open-source RMs, yielding a +17.0\% relative gain over its backbone in five-benchmark Overall accuracy. It matches or exceeds top-tier proprietary models on video and audio benchmarks, achieving 80.2\% on ShareGPT-Video, 66.8\% on Audio-HH, and 65.0\% on TA2T. Its efficacy is further validated through Best-of-$N$ inference and robust generalization to text-only preference tasks.
\end{itemize}

\section{Related Work}

\subsection{Multi-modal Large Language Models}
Open-source multimodal large language models have advanced rapidly recently, evolving from early vision-language instruction tuning (e.g., LLaVA) \cite{liu2023visualinstructiontuning,liu2024llavanext,li2024llavaonevision} to stronger billion-parameter backbones (e.g., InternVL) \cite{chen2023internvl,zhu2025internvl3}, and more native audio-visual processing in the Qwen family \cite{bai2023qwenvl,bai2025qwen25vl,xu2025qwen25omni}. Meanwhile, frontier models have begun to handle long-form, multi-turn multimodal reasoning \cite{feng2025video,yang2025r1}, and image/video understanding \cite{wang2026earlytom}. These advances substantially broaden the scope of multimodal applications, but also increase the demand for reliable and scalable alignment mechanisms that can generalize across modalities.

\subsection{Preference Alignment}
Preference alignment is commonly achieved via Reinforcement Learning from Human Feedback (RLHF) \cite{kaufmann2023rlhf}, with complementary directions including AI feedback (RLAIF) \cite{lee2023rlaif} and direct policy optimization methods such as DPO and GRPO \cite{rafailov2023dpo,shao2024deepseek}.
These paradigms typically rely on a \emph{reward model} to instantiate preferences as a learnable signal: It provides training-time supervision for post-training and can also serve as an inference-time reward score for reranking or Best-of-$N$ selection.
Extending preference optimization to multimodal settings is non-trivial due to heterogeneous inputs, modality-specific failure modes, and benchmark fragmentation \cite{sun2024aligning}.
Consequently, the reliability and coverage of the underlying RM often become the bottleneck in multimodal post-training pipelines, motivating reward supervision that is both scalable and interpretable.

\subsection{Reward Modeling}
\noindent\textbf{Multimodal Reward Models.}
Multimodal reward models (RMs) have lagged behind the generators they aim to align, with two recurring limitations.
First, modality coverage remains imbalanced and vision-centric: widely used preference resources and RM pipelines mainly target images and videos, supporting strong vision-oriented datasets, benchmarks, and reward models (e.g., ImageReward, Pick-a-Pic, and BaseReward)~\cite{xu2023imagereward,kirstain2023pickapic,zhang2025baserewardstrongbaselinemultimodal}, while audio remains less explored and standardized~\cite{sun2024aligning}.
Recent work such as Omni-Reward moves toward generalist omni-modal reward modeling by introducing multi-modality benchmarks and data, including audio, and training omni-modal RMs~\cite{jin2025omnireward}.
Second, interpretability remains limited: many RMs collapse judgments into scalar scores, while critic-style models provide free-form feedback that is hard to audit or control at the criterion level, as in LLaVA-Critic~\cite{xiong2025llava}.
To improve transparency, recent work incorporates Chain-of-Thought (CoT) reward reasoning in multimodal and text settings, such as UnifiedReward-Think, VR-Thinker, and RM-R1~\cite{wang2025unifiedrewardthink,wang2025vrthinker,chen2025rmr1}.
However, free-form traces do not enforce fixed criterion-wise accountability, and are often studied under restricted modality coverage or inconsistent evaluation interfaces.

\noindent\textbf{Rubric-grounded Reward Modeling.}
Rubric-grounded reward modeling constrains evaluation with explicit multi-criterion rubrics and dimension-wise justifications, improving controllability and auditability beyond scalar scores or unstructured critiques.
Prior work shows that criterion-level feedback can stabilize post-training and support targeted debugging in open-ended domains~\cite{gunjal2025rubrics}.
Recent studies further scale rubric supervision via automatic rubric construction and large rubric datasets~\cite{liu2026open,xie2026autorubric,li2026rubrichub}, and analyze how rubric design affects reliability and robustness under misspecification or spurious cues~\cite{zhang2026chasingtail,srivastava2025crome}.
Rubric-aware optimization and calibration have also been explored in post-training pipelines~\cite{zhou2026ruscarl,he2025rifl,jia2026openrs,xu2026positionbias}.
However, most rubric-grounded methods remain text-centric.
In contrast, Omni-RRM extends rubric-grounded reward modeling to multimodal alignment with a fixed five-criterion rubric, modality-aware instantiations, and a unified model that produces criterion-wise comparative judgments across text, image, video, and audio.
Our novelty does not lie in omni-modality, reasoning-style outputs, or RL refinement alone, but in unifying them through a shared rubric-grounded reward formulation, automatic rubric-grounded preference synthesis, and rubric-aware GRPO optimization.

\begin{figure*}[t!]
    \centering
    \includegraphics[width=\textwidth]{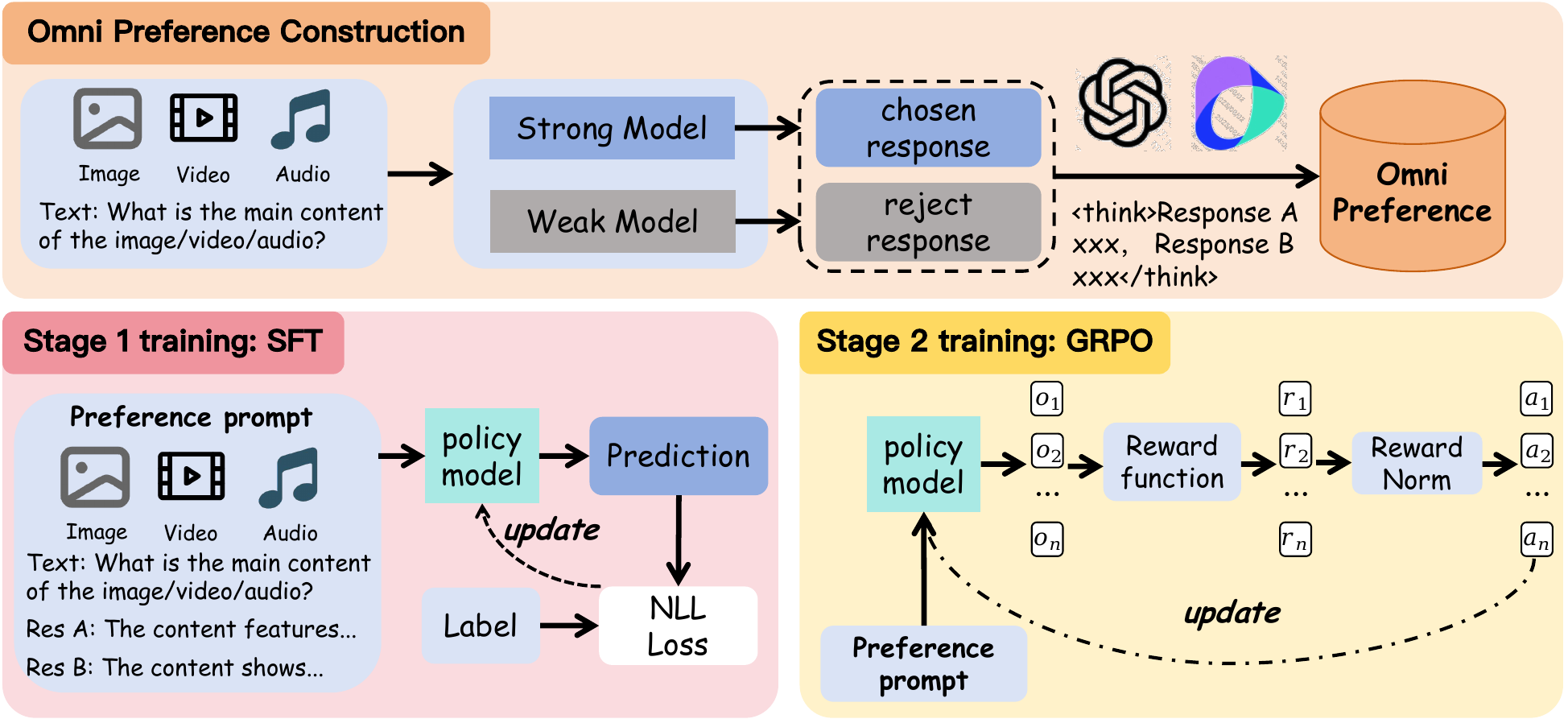}
    \caption
    { The overall pipeline for creating Omni-RRM. The process consists of two main phases. 
        \textbf{Top: Omni-Preference Construction.} Preference pairs are first automatically generated by contrasting outputs from a strong and a weak model. Powerful teacher models then annotate these pairs with rubric-grounded, multi-criteria justifications to create the final dataset. \textbf{Bottom: Two-Stage Training.} In Stage 1 (SFT), the policy model is trained on preference prompts using a standard NLL loss. In Stage 2 (GRPO), the model is further refined by generating multiple responses, scoring them with a rule-based reward function, and updating the policy based on the normalized reward. 
    }
    \label{fig:overall_pipeline}
\end{figure*}

\section{Method}

\paragraph{Problem setup and structured outputs.}
Given a multimodal context $x$ and a response pair $(y_A, y_B)$, Omni-RRM formulates reward modeling as structured generation rather than direct scalar preference prediction. It generates a rubric-grounded record
$r = (\hat{s}_A, \hat{s}_B, \hat{\ell}, \hat{\mathbf{j}}_{1:5}, \hat{\ell}')$, 
where $\hat{s}_A, \hat{s}_B \in [0, 10]$ are overall scores, $\hat{\ell} \in \{A, B, E\}$ is the preference label ($E$ denotes a tie), $\hat{\mathbf{j}}_{1:5}$ denotes five criterion-level comparative justifications, and $\hat{\ell}'$ is a redundant verdict copy for robust parsing. The five criteria---fluency/coherence, relevance, accuracy/completeness, reasoning quality, and safety/ethical alignment---provide a shared evaluation interface across text, image, video, and audio.

This structured format turns opaque scoring into a more auditable criterion-level judgment process. By requiring justifications before the final verdict, Omni-RRM encourages preferences grounded in human-inspectable evidence rather than holistic score patterns. The same schema is used in both SFT targets and GRPO reward computation, tying the rubric to optimization rather than only inference-time prompting.

\paragraph{Method overview.}
Our method comprises \textbf{Omni-Preference}, an automatically constructed preference dataset with teacher-reconciled rubric supervision (Table~\ref{tab:data-statistics-detailed}), and progressive training of \textbf{Omni-RRM} via SFT$\rightarrow$GRPO to sharpen discrimination on low-margin pairs (Figure~\ref{fig:overall_pipeline}).

\paragraph{Rubrics design.}
We adopt five broad rubrics to align with widely used axes in RLHF and reward-modeling practice and preference datasets.

\begin{itemize}
  \item \textbf{Fluency \& Coherence:} linguistic clarity and internal consistency of the response; consistent with reward-modeling datasets that explicitly annotate \emph{Coherence} as a fine-grained attribute (e.g., HelpSteer2)~\cite{wang2024helpsteer2}.
  \item \textbf{Relevance:} adherence to the user request and provided context, reflecting the \emph{helpfulness} / instruction-following emphasis in RLHF preference data and instruction-tuning with human feedback (e.g., HH-RLHF; InstructGPT)~\cite{bai2022helpful_harmless,ouyang2022training}.
  \item \textbf{Accuracy \& Completeness:} factual correctness and sufficient coverage of key requested information, aligning with \emph{Correctness} (and related truthfulness/correctness objectives) commonly tracked in RM datasets and instruction-following RLHF evaluations (e.g., HelpSteer2; InstructGPT)~\cite{wang2024helpsteer2,ouyang2022training}.
  \item \textbf{Reasoning Quality:} soundness of inference and absence of contradictions, introduced to make judgments actionable beyond surface-level correctness and to complement consistency-focused analyses in preference modeling~\cite{bai2022helpful_harmless,ouyang2022training}.
  \item \textbf{Safety \& Ethical Alignment:} avoidance of harmful or unsafe content, matching the ``helpful vs.\ harmless'' framing and explicit safety-alignment formulations in HH-RLHF and Safe RLHF, as well as safety/toxicity-focused RLHF evaluation goals~\cite{bai2022helpful_harmless,dai2023saferlhf,ouyang2022training}.
\end{itemize}

\paragraph{Modality-aware instantiation.}
Although the five rubrics are shared across modalities, the evidence used to instantiate each criterion is modality-specific. For image inputs, accuracy and reasoning emphasize visual grounding, object attributes, spatial relations, and OCR-sensitive details. For video inputs, the same criteria additionally consider temporal ordering, action progression, and event consistency. For audio inputs, they focus on acoustic event recognition, spoken content, and consistency between audio cues and textual responses. This design keeps the evaluation interface unified while allowing criterion-level evidence to reflect modality-specific failure modes.

\subsection{Omni-Preference: A Rubric-Grounded Dataset for Omni-Modal Reward Modeling}
We construct Omni-Preference, a fully automated dataset synthesized via our rubric-grounded preference synthesis pipeline. It is designed to provide high-fidelity reward signals across image, video, and audio. Unlike traditional preference sets, each sample in Omni-Preference is anchored by a multidimensional record $(x, y_A, y_B)$, annotated with teacher-reconciled preferences and structured rationales (Table~\ref{tab:data-statistics-detailed}).

\noindent\textbf{Stage 1: Diverse Candidate Pair Generation.}
To populate the reward model's training distribution, we generate candidate pairs via a capability-gap strategy. For each context $x$ sampled from omni-modal datasets, we sample responses from a strong generator $M_S$ and a weak generator $M_W$:
\begin{equation}
(y_A, y_B) = \big(M_S(x), M_W(x)\big).
\end{equation}
This ensures a wide coverage of quality margins. All definitive reward labels are assigned via the rubric-grounded consensus mechanism in Stage 2 to handle potential model reversals. We empirically compare against single-model rollouts in Supplementary Sec.~B.3. This comparison shows that single-model rollouts are frequently judged as ties and yield very few large-margin pairs, making them inefficient for constructing high-confidence preference supervision under a fixed budget. In contrast, capability-gap pairing increases response diversity while leaving all final labels to the teacher-reconciled Stage-2 consensus.

\noindent\textbf{Stage 2: Rubric-Grounded Annotation and Reconciliation.}
To ensure the precision of the reward signal, we employ two heterogeneous teachers (GPT-4o-mini and Gemini-2.0-Flash). Each teacher provides: (i) scalar scores, (ii) a categorical verdict $\ell \in \{A, B, E\}$, and (iii) a rubric-grounded justification across five shared criteria. 

We apply a strict consensus filter: a pair is retained only if both teachers reach an identical non-tie verdict that is numerically consistent with their score ranking and rubric-grounded justification. This dual-teacher reconciliation minimizes individual bias and reinforces the density of the rubric-grounded justification. We quantify sample difficulty using the reconciled score margin $\Delta = |s_A - s_B|$, where $\Delta < 2$ defines Hard pairs critical for refining the reward model's discriminative boundary. Details of teacher prompting, reconciliation, and filtering rules are in Supplementary Sec.~B.1.

\noindent\textbf{Data scale and signal quality.}
We prioritize supervision reliability and cross-modality coverage over raw scale. This reliability-first design is particularly important for low-margin preference pairs, where noisy labels can easily blur the reward boundary and destabilize subsequent RL refinement. Detailed reliability analyses of synthetic preferences, including human validation and train--test contamination checks, are provided in Supplementary Sec.~B.5.

\begin{table*}[t!]
  \centering
  \footnotesize 
  \caption{Statistics of the synthesized \textbf{Omni-Preference} dataset.}
  \label{tab:data-statistics-detailed}

  \begingroup
  \setlength{\tabcolsep}{3pt}
  \renewcommand{\arraystretch}{1.15}
  \newcolumntype{L}{>{\raggedright\arraybackslash}m{4.5em}}
  \newcolumntype{R}{>{\raggedleft\arraybackslash}m{3.5em}}  

  \begin{tabularx}{\textwidth}{@{} L >{\raggedright\arraybackslash\hsize=0.65\hsize}X >{\centering\arraybackslash\hsize=1.2\hsize}X >{\centering\arraybackslash\hsize=1.2\hsize}X R @{}}
    \toprule
    \textbf{Modality} & \textbf{Data Source} & \textbf{Strong Model ($M_S$)} & \textbf{Weak Model ($M_W$)} & \textbf{Samples} \\
    \midrule

    \multirow{2}{*}{Image} & \multirow{2}{*}{RLAIF-V} & Qwen2.5-VL-7B & Qwen2.5-VL-3B & \multirow{2}{*}{17.0k} \\
    & & Qwen2.5-VL-7B & LLaVA-1.5-7B & \\
    \midrule

    Video & video datasets\textsuperscript{\dag} & Qwen2.5-VL-7B & Qwen2.5-VL-3B & 12.2k \\
    \midrule

    \multirow{3}{*}{Audio} & \multirow{3}{*}{Clotho-AQA} & R1-AQA-7B & Qwen2-Audio-7B & \multirow{3}{*}{11.8k} \\
    & & Qwen2.5-Omni-7B & Qwen2-Audio-7B & \\
    & & Qwen2.5-Omni-7B & Qwen2.5-Omni-3B & \\
    \midrule[1pt]

    \textbf{Total} & \multicolumn{3}{l}{} & \textbf{41.0k} \\
    \bottomrule
  \end{tabularx}

  \vspace{0.2em}
  \textsuperscript{\dag}ActivityNet, Charades, Ego4D, NextQA, and YouCook2.
  \endgroup
\end{table*}

\subsection{Training Omni-RRM: A Progressive SFT--RL Approach}
We train Omni-RRM with a progressive two-stage strategy: supervised fine-tuning (SFT) to learn rubric-grounded structured outputs, followed by reinforcement learning to refine discrimination on low-margin preference pairs.

The RL stage uses a bounded composite objective that combines: (i) preference correctness, (ii) the quality of the rubric-grounded justification, and (iii) a lightweight schema-validity guardrail to prevent format collapse during exploration.
Such schema constraints are commonly used to stabilize structured critics under RL \cite{wang2024helpsteer2,lu2025schemarl}.

\paragraph{Stage 1: Supervised Fine-Tuning (SFT).}
SFT teaches the model to follow the output interface and generate coherent rubric-grounded justifications. Given input $(x,y_A,y_B)$ and a target structured record represented as a token sequence $\mathbf{o}_{1:T}$, we optimize the standard negative log-likelihood loss:
\begin{equation}
\mathcal{L}_{\text{SFT}}(\theta) = - \sum_{i=1}^{T} \log P_\theta(o_i \mid \mathbf{o}_{<i}, x, y_A, y_B).
\end{equation}
We use lightweight LoRA fine-tuning \cite{hu2022lora} for efficiency. By the end of SFT, the model reliably produces schema-compliant outputs with non-trivial rubric-grounded justifications, providing a stable initialization for RL.

\paragraph{Stage 2: Reinforcement Learning with GRPO.}
SFT provides a schema-stable cold start so that RL updates can focus on improving low-margin preference discrimination rather than repairing output format \cite{guo2025deepseek,lu2025schemarl}.
We then refine Omni-RRM with Group Relative Policy Optimization (GRPO), following the group-sampling and relative-advantage optimization recipe popularized in DeepSeek-R1-style training \cite{shao2024deepseek,guo2025deepseek}. For each input $(x,y_A,y_B)$, we sample a group of $k$ structured outputs and assign each output a bounded composite reward:
\begin{equation}
R = w_{\text{fmt}} R_{\text{fmt}} + w_{\text{pref}} R_{\text{pref}} + w_{\text{rub}} R_{\text{rub}}.
\end{equation}
$R_{\text{fmt}}$ enforces strict JSON/schema validity and required fields (e.g., overall scores $\hat{s}_A/\hat{s}_B$, the categorical verdict, and the rubric-grounded justification fields).
$R_{\text{pref}}$ rewards correct preference prediction and score--verdict consistency, i.e., the predicted verdict $\hat{\ell}\in\{A,B,E\}$ should agree with the score ordering (e.g., $\hat{\ell}=A\Rightarrow \hat{s}_A>\hat{s}_B$, $\hat{\ell}=B\Rightarrow \hat{s}_B>\hat{s}_A$, and $\hat{\ell}=E\Rightarrow \hat{s}_A=\hat{s}_B$).
$R_{\text{rub}}$ encourages substantive, rubric-grounded justifications, serving as the structural regularization term that compels the reward model to derive its verdict through a disciplined, criterion-based analysis; it is computed using justification-consistency heuristics and consistency checks (details below).
We optimize the policy using group-normalized advantages and a clipped policy-gradient objective with a KL penalty to the SFT reference policy. Full GRPO update equations, reward rules, and weights are provided in Supplementary Sec.~A.1.

Overall, the progressive SFT--GRPO regimen stabilizes the rubric-grounded output interface and improves discrimination on low-margin preference pairs. A detailed compute budget for data synthesis, teacher annotation, training, and evaluation is provided in Supplementary Sec.~A.2.

\section{Experiments}

\subsection{Benchmarks and Metrics}
We evaluate Omni-RRM on preference benchmarks spanning image, video, and audio. 
Our primary metric is Preference Accuracy (\%), i.e., the fraction of test pairs where the predicted winner $\hat{\ell}\in\{A,B\}$ matches the benchmark's ground-truth preference under its official protocol.

\noindent\textbf{Image.} We report results on VL-RewardBench (VL-Reward)~\cite{Li2025vlreward} using its recommended \emph{Reasoning} split, and on MM-RewardBench~\cite{yasunaga2025multimodal} using its primary preference accuracy metric.

\noindent\textbf{Video.} We derive a pairwise preference test set from ShareGPT-Video DPO (ShareGPT-Video)~\cite{zhang2024direct} by retaining pairs with sufficiently large human score margins to obtain clear winners.

\noindent\textbf{Audio.}
We use two audio-conditioned preference benchmarks. Audio-HH-RLHF (Audio-HH) is constructed by applying TTS to prompts from HH-RLHF~\cite{bai2022helpful_harmless}, while retaining the original human-written chosen/rejected responses and preference labels. This yields a controlled audio-conditioned preference setting, though it does not cover real-world acoustic noise or speaker variability. We further evaluate on Omni-RewardBench-TA2T (TA2T)~\cite{jin2025omnireward}, a public audio-conditioned reward benchmark, to assess external audio generalization.

\begin{table}[t!]
  \centering
  \small
  \caption{Main results on multi-modal preference benchmarks. We report accuracy (\%) across benchmarks. The \textbf{best} in each column is bold and the \underline{second best} is underlined. `---' indicates that the model is not evaluated on the corresponding benchmark. \textbf{Overall} is the mean accuracy over the five benchmark columns. TA2T denotes Omni-RewardBench-TA2T with tie cases included; full with-/without-tie results are provided in Supplementary Sec.~C.6. Values in parentheses (ours) are relative improvements (\%) over the corresponding Qwen2.5-Omni backbones.}
  \label{tab:main_results}

  \begingroup
  \setlength{\tabcolsep}{2.0pt}
  \renewcommand{\arraystretch}{1.10}

  \resizebox{\columnwidth}{!}{%
  \begin{tabular}{@{}l c c c c c c@{}}
    \toprule
    \textbf{Model} &
    \makecell[c]{\textbf{VL-}\\ \textbf{Reward}} &
    \makecell[c]{\textbf{MM-}\\ \textbf{RewardBench}} &
    \makecell[c]{\textbf{ShareGPT-}\\ \textbf{Video}} &
    \makecell[c]{\textbf{Audio-}\\ \textbf{HH}} &
    \textbf{TA2T} &
    \textbf{Overall} \\
    \midrule

    \multicolumn{7}{@{}l}{\textit{Proprietary models}} \\
    \midrule
    GPT-4o-mini             & 59.8 & 61.9 & 53.9 & 58.2 & 57.9 & 58.3 \\
    Doubao-1.5-Vision-Pro   & \underline{77.3} & 68.0 & 77.0 & ---  & ---  & --- \\
    Gemini-2.0-Flash        & 73.4 & 62.8 & 74.6 & 60.1 & 59.9 & 66.2 \\
    Gemini-2.5-Pro          & \textbf{79.6} & 63.3 & \underline{78.8} & \underline{66.5} & \underline{64.9} & \textbf{70.6} \\
    \midrule

    \multicolumn{7}{@{}l}{\textit{Open-source models}} \\
    \midrule
    Qwen2.5-Omni-3B         & 53.7 & 53.9 & 58.1 & 58.7 & 47.9 & 54.5 \\
    Qwen2.5-Omni-7B         & 57.8 & 57.5 & 66.3 & 62.4 & 56.9 & 60.2 \\
    Qwen2.5-VL-3B           & 53.2 & 53.3 & 61.2 & ---  & ---  & --- \\
    Qwen2.5-VL-7B           & 58.2 & 56.0 & 70.5 & ---  & ---  & --- \\
    Qwen2.5-VL-72B          & 62.3 & 63.5 & 72.9 & ---  & ---  & --- \\
    \midrule

    \multicolumn{7}{@{}l}{\textit{Open-source reward models}} \\
    \midrule
    LLaVA-Critic-7B         & 54.1 & 56.0 & ---  & ---  & ---  & --- \\
    Skywork-VL-Reward-7B    & 60.4 & 67.4 & 59.9 & ---  & ---  & --- \\
    UnifiedReward-think-7B  & 66.6 & 71.4 & 77.8 & ---  & ---  & --- \\
    Omni-RewardModel-BT     & 60.4 & 58.4 & 63.7 & 61.3 & 60.5 & 60.9 \\
    R1-Reward-7B            & 65.8 & \underline{72.3} & 58.7 & ---  & ---  & --- \\
    \midrule

    \multicolumn{7}{@{}l}{\textbf{Ours}} \\
    \midrule
    Omni-RRM-3B (\textit{sft})    & 56.8 & 58.1 & 64.9 & 60.3 & 54.3 & 58.9 \\
    Omni-RRM-7B (\textit{sft})    & 60.4 & 61.0 & 70.5 & 62.8 & 58.5 & 62.6 \\

    Omni-RRM-3B (\textit{sft+rl}) &
      \makecell[c]{58.5\\{\scriptsize(+8.9\%)}} &
      \makecell[c]{68.9\\{\scriptsize(+27.8\%)}} &
      \makecell[c]{67.4\\{\scriptsize(+16.0\%)}} &
      \makecell[c]{65.1\\{\scriptsize(+10.9\%)}} &
      \makecell[c]{61.1\\{\scriptsize(+27.5\%)}} &
      \makecell[c]{64.2\\{\scriptsize(+17.9\%)}} \\

    Omni-RRM-7B (\textit{sft+rl}) &
      \makecell[c]{67.1\\{\scriptsize(+16.1\%)}} &
      \makecell[c]{\textbf{72.9}\\{\scriptsize(+26.8\%)}} &
      \makecell[c]{\textbf{80.2}\\{\scriptsize(+21.0\%)}} &
      \makecell[c]{\textbf{66.8}\\{\scriptsize(+7.1\%)}} &
      \makecell[c]{\textbf{65.0}\\{\scriptsize(+14.3\%)}} &
      \makecell[c]{\underline{70.4}\\{\scriptsize(+17.0\%)}} \\
    \bottomrule
  \end{tabular}%
  }
  \endgroup
\end{table}

\subsection{Baselines}

We compare Omni-RRM against baselines spanning proprietary MLLMs, open-source MLLMs, and open-source multimodal reward models. 
\textbf{Proprietary models} (e.g., GPT-4o-mini~\cite{openai2024gpt4o} and Gemini-2.0-Flash~\cite{google2025gemini}) serve as strong reference points.
\textbf{Open-source MLLMs} are represented by the Qwen2.5 family~\cite{xu2025qwen25omni}; we include the original Qwen2.5-Omni backbones as \emph{unaligned} baselines to quantify gains from our post-training.
We further compare against \textbf{open-source reward models} trained for multimodal preference judgment, including Omni-Reward~\cite{jin2025omnireward}, LLaVA-Critic-7B~\cite{xiong2025llava}, R1-Reward-7B~\cite{zhao2025r1reward}, and UnifiedReward-Think-7B~\cite{wang2025unifiedrewardthink}.
To isolate the effect of rubric-structured supervision and rubric-aware RL, we include two matched variants under the same SFT+RL budget:
\textbf{Omni-RRM (w/o rationale)}, which removes five-criterion rubric rationales during SFT+RL, and
\textbf{Omni-RRM (w/o rubric RL)}, which ablates the rubric-related reward term during GRPO.

\subsection{Experimental Results}
\label{sec:exp_results}

Our main results in Table~\ref{tab:main_results} yield several primary insights.
Additional analyses on Hard/Easy subsets and text-only transfer are provided in Supplementary Secs.~C.1--C.2.

\paragraph{Omni-RRM achieves state-of-the-art omni-modal performance.}
Omni-RRM-7B sets a new state-of-the-art among open-source reward models on video and audio benchmarks, achieving 80.2\% on ShareGPT-Video, 66.8\% on Audio-HH-RLHF (Audio-HH), and 65.0\% on TA2T. Its five-benchmark Overall accuracy of 70.4\% closely matches Gemini-2.5-Pro (70.6\%), despite being an order of magnitude smaller. On image benchmarks, it also achieves the highest score on MM-RewardBench (\textbf{72.9\%}), outperforming both specialized reward models and general-purpose proprietary MLLMs. Including TA2T in the main evaluation ensures that our audio conclusion is not tied to a single controlled benchmark. The rankings on Audio-HH and TA2T are highly correlated across common models, supporting Audio-HH as a meaningful controlled audio proxy; detailed with-/without-tie results and correlation analysis are provided in Supplementary Sec.~C.6.

\paragraph{Rubric-grounded rationales are vital for high-fidelity reward modeling.}
To isolate the impact of \textbf{rubric-grounded} supervision, we compare against Omni-RRM (w/o rationale), a variant trained without generating rationales (Table~\ref{tab:ablations}). The sharp decline in this variant indicates that rubric-grounded rationales are not merely auxiliary text, but provide useful structural regularization for preference discrimination. By requiring criterion-level evidence before the final verdict, the model becomes less reliant on holistic score patterns or shortcut multimodal cues. We further ablate rubric-aware reinforcement via Omni-RRM (w/o rubric RL), which removes the $R_{\text{rub}}$ term during GRPO; this variant likewise underperforms the full model in the four-benchmark ablation suite (68.1\% vs.\ 71.8\% Overall for 7B). Together, these ablations show that rubric-grounded rationales and rubric-aware optimization jointly improve omni-modal reward modeling.

\paragraph{Progressive SFT+GRPO training drives substantial performance gains.}
Our two-stage training regimen yields significant improvements over both the unaligned backbone and the SFT-only baseline. While SFT establishes a foundational capability for \textbf{reward modeling}, the subsequent \textbf{GRPO refinement} further sharpens the reward boundary. For example, Omni-RRM-7B improves from \textbf{62.6\%} (SFT) to \textbf{70.4\%} (SFT+RL) in five-benchmark Overall accuracy, with particularly strong gains on MM-RewardBench (\textbf{61.0\% $\rightarrow$ 72.9\%}), ShareGPT-Video (\textbf{70.5\% $\rightarrow$ 80.2\%}), and TA2T (\textbf{58.5\% $\rightarrow$ 65.0\%}). Compared to the original Qwen2.5-Omni-7B backbone, our full model achieves a \textbf{+17.0\% relative gain} in overall accuracy, validating that RL-based optimization effectively improves reward calibration. This improvement is especially pronounced on low-contrast examples: difficulty-stratified evaluation in Supplementary Sec.~C.1 shows that Omni-RRM yields larger average gains on Hard pairs than Easy pairs (+12.0 pp vs. +6.6 pp), supporting our design focus on low-margin preference discrimination.

\paragraph{Specialized reward modeling outweighs raw model scale.}
We find that \textbf{reward modeling} is a specialized capability that does not scale linearly with general pre-training: reward models explicitly optimized for preference supervision can significantly outperform much larger general-purpose MLLMs. For example, on MM-RewardBench, the proprietary Gemini-2.5-Pro (\textbf{63.3\%}) lags substantially behind specialized open-source reward models, including our Omni-RRM-7B (\textbf{72.9\%}), R1-Reward-7B (\textbf{72.3\%}), and UnifiedReward-Think-7B (\textbf{71.4\%}). This performance gap suggests that high-fidelity reward modeling depends critically on \textbf{targeted supervision} and \textbf{rubric-grounded post-training objectives} rather than sheer parameter count or pre-training scale alone.

\begin{table}[t]
  \centering
  \small
  \caption{Ablations on rubric-grounded supervision and rubric-aware optimization. \textbf{w/o rationale} removes the five-criterion rubric rationales during SFT+RL, and \textbf{w/o rubric RL} removes the rubric-related reward term during GRPO. Overall averages the four ablation benchmarks.}
  \label{tab:ablations}

  \begingroup
  \setlength{\tabcolsep}{2.4pt}
  \renewcommand{\arraystretch}{1.10}

  \resizebox{\columnwidth}{!}{%
  \begin{tabular}{@{}p{0.50\columnwidth} c c c c c@{}}
    \toprule
    \textbf{Variant} &
    \makecell[c]{\textbf{VL-}\\ \textbf{Reward}} &
    \makecell[c]{\textbf{MM-}\\ \textbf{RewardBench}} &
    \makecell[c]{\textbf{ShareGPT-}\\ \textbf{Video}} &
    \makecell[c]{\textbf{Audio-}\\ \textbf{HH}} &
    \textbf{Overall} \\
    \midrule

    \multicolumn{6}{@{}l}{\textbf{Omni-RRM-3B}} \\
    \midrule
    \quad w/o rationale   & 56.9 & 54.7 & 60.0 & 63.8 & 58.9 \\
    \quad w/o rubric RL   & 57.4 & 64.1 & 65.6 & 64.6 & 62.9 \\
    \quad full model      & 58.5 & 68.9 & 67.4 & 65.1 & 65.0 \\
    \midrule

    \multicolumn{6}{@{}l}{\textbf{Omni-RRM-7B}} \\
    \midrule
    \quad w/o rationale   & 64.0 & 60.6 & 68.7 & 64.3 & 64.4 \\
    \quad w/o rubric RL   & 65.2 & 66.6 & 75.4 & 65.0 & 68.1 \\
    \quad full model      & 67.1 & 72.9 & 80.2 & 66.8 & 71.8 \\
    \bottomrule
  \end{tabular}%
  }
  \endgroup
\end{table}

\subsection{Inference-Time Alignment via Best-of-$N$}
\label{sec:bon}

To validate the practical utility of \textbf{Omni-RRM} as a plug-and-play alignment tool, we employ a Best-of-$N$ (BoN) re-ranking protocol using a fixed base generator. Specifically, we use \textbf{Qwen2.5-Omni-7B} to sample $N$ candidate responses per query and select the final output via two distinct strategies:
(i) \textbf{Self-consistency@N}, which performs a majority vote over the multiple-choice options among the $N$ candidates; and
(ii) \textbf{BoN@N}, which selects the optimal candidate based on an external \textbf{reward model}. 
For the latter, we implement a \textbf{single-elimination tournament}: the reward model performs pairwise comparisons under a fixed bracket order, with the preferred response advancing until a single winner remains ($N{-}1$ comparisons per query). This setup isolates the effect of \textbf{inference-time alignment} from any changes to the generator's underlying parameters.

We evaluate BoN performance on three representative benchmarks: \textbf{MMMU} (Image)~\cite{yue2024mmmu}, \textbf{Video-MME} (Video)~\cite{fu2025video}, and \textbf{AVQA} (Audio-only setting)~\cite{yang2022avqa}. We compare five open-source \textbf{reward models}: Qwen2.5-Omni-7B, Qwen2.5-VL-7B, LLaVA-Critic-7B, UnifiedReward-think-7B, and our \textbf{Omni-RRM-7B}. Results are reported only for modalities supported by each respective model, with $N \in \{2,4,6,8,10\}$.

\begin{figure}[t!]
    \centering
    \includegraphics[width=1\textwidth]{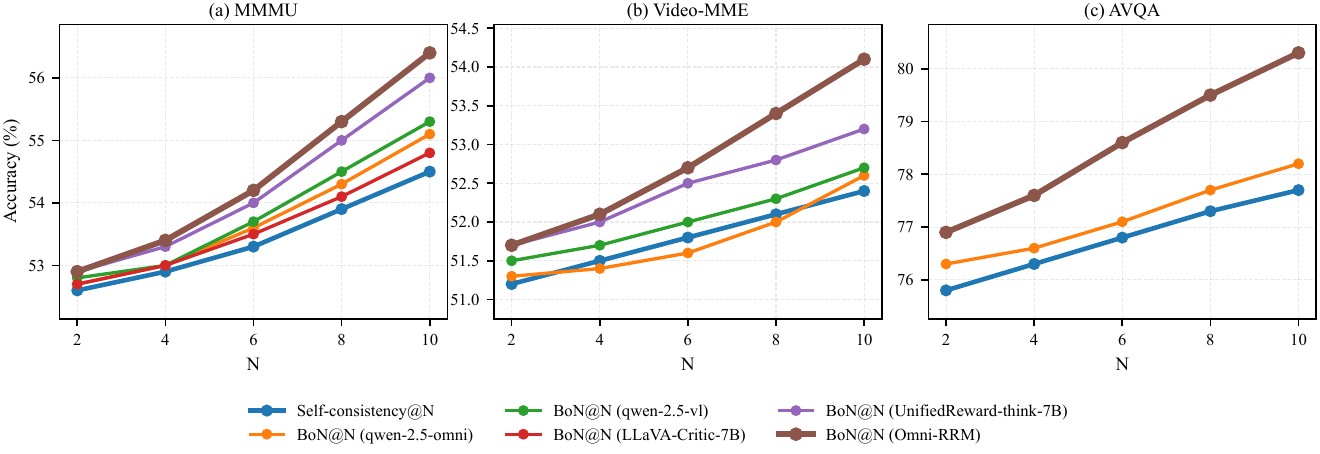}
    \caption{
        Best-of-$N$ inference-time alignment across varying $N$ and reward models.
        Using Qwen2.5-Omni-7B as the base generator, we evaluate Self-consistency@N (majority vote) against BoN@N (single-elimination tournament) on MMMU (Image), Video-MME (Video), and AVQA (Audio).
    }
    \label{fig:bon_curve}
\end{figure}

As illustrated in Figure~\ref{fig:bon_curve}, the choice of \textbf{reward model} significantly influences BoN effectiveness, with the performance gap over self-consistency typically widening as $N$ increases. Across all modalities, \textbf{Omni-RRM-7B} consistently yields the most substantial improvements. At $N{=}10$, Omni-RRM improves upon Self-consistency by \textbf{+1.9} on MMMU ($54.5 \rightarrow 56.4$), \textbf{+1.7} on Video-MME ($52.4 \rightarrow 54.1$), and \textbf{+2.6} on AVQA (\textbf{audio-only}; $77.7 \rightarrow 80.3$). 

Although Omni-RRM introduces extra RM decoding latency due to structured rationales, it delivers stronger BoN selection gains on video and audio tasks, suggesting that criterion-level reasoning is most beneficial when fine-grained discrimination among diverse candidates is required.

These results underscore the scalability of our \textbf{rubric-grounded reward modeling} in complex inference-time scenarios. As the candidate pool $N$ expands, the base generator yields a \textbf{more diverse set of potential responses}, increasing the likelihood that the pool contains high-quality candidates that surpass standard greedy decoding. However, identifying these optimal responses among a larger set of plausible alternatives requires \textbf{enhanced discrimination granularity}. \textbf{Omni-RRM} thrives in this high-$N$ regime by leveraging its structured rubric rationales to conduct exhaustive, multi-dimensional comparisons. This systematic evaluation allows the model to effectively \textbf{capture subtle quality gains} and consistently \textbf{surface the best-performing responses} from the expanded search space, leading to the superior performance observed across all three modalities. Detailed numerical results and a latency analysis are provided in Supplementary Secs.~C.3--C.4.

\subsection{Inter-Modality Dynamics in Omni-Modal Reward Modeling}
\label{sec:ablation_omni}

A fundamental challenge in omni-modal alignment lies in understanding how preference knowledge is encoded and transferred across heterogeneous sensory inputs. Rather than viewing reward modeling as a collection of disjoint tasks, we treat it as a unified \textbf{omni-modal reasoning problem} anchored by a shared rubric. By systematically varying the \textbf{supervision mixture} of image, video, and audio data, we aim to map the \textbf{omni-modal dynamics} and quantify how preference signals in one domain facilitate or reinforce the reward boundaries in others. This investigation moves beyond simple coverage to explore whether a \textbf{unified rubric interface} can effectively harmonize diverse omni-modal inputs into a consistent, modality-agnostic preference space.

\paragraph{Experimental setup.}
All ablations utilize the same 3B backbone and identical SFT+GRPO hyperparameters. We maintain a constant total training budget (in terms of updates and pairs) across all settings by subsampling, ensuring that performance variance is strictly driven by the \textbf{composition of omni-modal supervision} rather than raw data volume. We evaluate preference accuracy (\%) across \textbf{VL-RewardBench} (Image), \textbf{ShareGPT-Video} (Video), and \textbf{Audio-HH-RLHF} (Audio).

\begin{figure}[t!]
    \centering
    \includegraphics[width=0.75\linewidth]{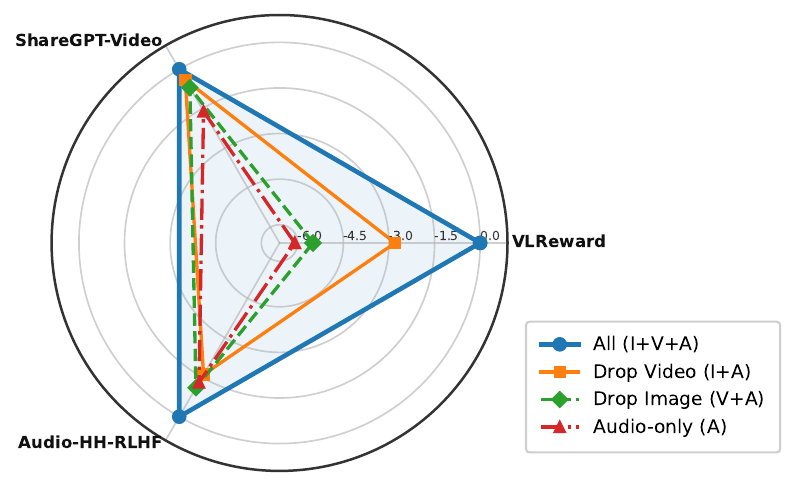}
    \caption{
        \textbf{Exploring omni-modal synergy and cross-modal transfer in reward modeling.}
        All variants utilize the same 3B backbone and identical SFT+GRPO pipeline, with only the supervision modality composition being varied (All / Drop Video / Drop Image / Audio-only). Each spoke represents an evaluation benchmark (VL-RewardBench / ShareGPT-Video / Audio-HH-RLHF), while each ring plots the performance of a specific training subset.
    }
    \label{fig:omni_ablation}
\end{figure}

\paragraph{Results and Analysis.}
Figure~\ref{fig:omni_ablation} reveals a complex landscape of \textbf{omni-modal synergy} and \textbf{reciprocal reinforcement}:

\noindent\textbf{(1) Global convergence through omni-modal joint training.} 
The full-modality configuration (I+V+A) achieves a \textbf{superior performance envelope} that transcends any modality-specific or partially combined variant. This demonstrates that joint training across all three modalities does not lead to optimization conflicts; instead, the diversity of omni-modal inputs facilitates a more robust and \textbf{generalized convergence} on the underlying rubric criteria, outperforming models trained on narrower distributions.

\noindent\textbf{(2) Reciprocal dependency and semantic anchoring.} 
We observe a \textbf{collateral performance decline} whenever any single modality is excluded, suggesting that preference knowledge is not siloed within specific domains. For instance, removing video-level reasoning patterns (\textbf{I+A}) not only eliminates video capabilities but also compromises the model's judgment on image and audio tasks.  This confirms that our \textbf{rubric-grounded rationales} act as a shared semantic anchor: the "logic" learned in one modality serves as a reference for others, creating a mutually reinforcing omni-modal reward landscape.

\noindent\textbf{(3) Asymmetric gains in low-resource omni-modal domains.} 
A pivotal finding is the \textbf{asymmetric performance boost} in the audio domain. Omni-modal training (\textbf{65.1\%}) significantly outperforms audio-only training (\textbf{63.8\%}) under an identical budget. This highlights that the high-granularity preference structures inherited from vision and video datasets provide a \textbf{structural scaffold} for audio. By "borrowing" the well-defined boundaries from high-resource modalities, \textbf{Omni-RRM} effectively overcomes the inherent scarcity of audio preference data, validating the strategic efficiency of unified omni-modal reward modeling.

\section{Conclusion}
\label{sec:conclusion}

In this paper, we presented \textbf{Omni-RRM}, a rubric-grounded framework for omni-modal reward modeling. Our approach alleviates the scalability and consistency bottlenecks of manual labeling through automatic rubric-grounded preference synthesis, yielding the \textbf{Omni-Preference} dataset enriched with dense criterion-level supervision signals. By shifting from coarse, black-box scoring to structured five-criterion justifications, Omni-RRM introduces structural regularization that makes reward judgments more auditable and less dependent on opaque scalar patterns or shortcut cues across image, video, and audio. Extensive evaluations demonstrate that \textbf{Omni-RRM} achieves state-of-the-art accuracy among open-source reward models, reaching 70.4\% five-benchmark Overall accuracy and closely matching strong proprietary MLLMs on challenging multimodal benchmarks. Our analysis of omni-modal dynamics further suggests that a unified rubric-grounded framework can induce cross-modal transfer, allowing preference signals from heterogeneous modalities to reinforce a more consistent omni-modal reward boundary. Furthermore, we validate its utility as a plug-and-play alignment tool via Best-of-$N$ inference, demonstrating its practical effectiveness for inference-time response selection. By harmonizing omni-modal coverage with structured transparency, this work provides a robust foundation for building more trustworthy, scalable, and efficient human-AI alignment systems.

\section*{Acknowledgements}

This work was supported in part by the National Natural Science Foundation of China under Grant Nos. 62576046, 62301066, 62406028, and 62476224; the Beijing Academy of Artificial Intelligence under Grant No. Z251100008125041; the Key Project of Philosophy and Social Sciences Research, Ministry of Education of China, ``Research on Risk Governance and Capability Building of Generative Artificial Intelligence Systems,'' under Grant No. 24JZD040; the Graduate Education and Teaching Reform Research Project of Beijing University of Posts and Telecommunications under Grant No. 2025YZ010; and the Beijing Natural Science Foundation under Grant No. L259043.

%
%
\bibliographystyle{splncs04}
\bibliography{main}

\begin{thebibliography}{10}
\providecommand{\url}[1]{\texttt{#1}}
\providecommand{\urlprefix}{URL }
\providecommand{\doi}[1]{https://doi.org/#1}

\bibitem{bai2023qwenvl}
Bai, J., Bai, S., Chu, Y., Cui, Z., Dang, K., Deng, X., Fan, Y., Ge, W., Han,
  Y., Huang, F., Hui, B., Ji, L., Li, M., Lin, J., Lin, R., Liu, D., Liu, G.,
  Lu, C., Lu, K., Ma, J., Men, R., Ren, X., Ren, X., Tan, C., Tan, S., Tu, J.,
  Wang, P., Wang, S., Wang, W., Wu, S., Xu, B., Xu, J., Yang, A., Yang, H.,
  Yang, J., Yang, S., Yao, Y., Yu, B., Yuan, H., Yuan, Z., Zhang, J., Zhang,
  X., Zhang, Y., Zhang, Z., Zhou, C., Zhou, J., Zhou, X., Zhu, T.: Qwen
  technical report (2023), \url{https://arxiv.org/abs/2309.16609}, accessed: 26
  Jun 2026

\bibitem{bai2025qwen25vl}
Bai, S., Chen, K., Liu, X., Wang, J., Ge, W., Song, S., Dang, K., Wang, P.,
  Wang, S., Tang, J., Zhong, H., Zhu, Y., Yang, M., Li, Z., Wan, J., Wang, P.,
  Ding, W., Fu, Z., Xu, Y., Ye, J., Zhang, X., Xie, T., Cheng, Z., Zhang, H.,
  Yang, Z., Xu, H., Lin, J.: Qwen2.5-vl technical report (2025),
  \url{https://arxiv.org/abs/2502.13923}, accessed: 26 Jun 2026

\bibitem{bai2022helpful_harmless}
Bai, Y., Jones, A., Ndousse, K., Askell, A., Chen, A., DasSarma, N., Drain, D.,
  Fort, S., Ganguli, D., Henighan, T., Joseph, N., Kadavath, S., Kernion, J.,
  Conerly, T., El-Showk, S., Elhage, N., Hatfield-Dodds, Z., Hernandez, D.,
  Hume, T., Johnston, S., Kravec, S., Lovitt, L., Nanda, N., Olsson, C.,
  Amodei, D., Brown, T.B., Clark, J., McCandlish, S., Olah, C., Mann, B.,
  Kaplan, J.: Training a helpful and harmless assistant with reinforcement
  learning from human feedback. arXiv preprint  \textbf{2204.05862} (2022),
  \url{https://arxiv.org/abs/2204.05862}, accessed: 26 Jun 2026

\bibitem{chen2025rmr1}
Chen, X., Li, G., Wang, Z., Jin, B., Qian, C., Wang, Y., Wang, H., Zhang, Y.,
  Zhang, D., Zhang, T., Tong, H., Ji, H.: Rm-r1: Reward modeling as reasoning.
  arXiv preprint arXiv:2505.02387  (2025),
  \url{https://arxiv.org/abs/2505.02387}, accessed: 26 Jun 2026

\bibitem{chen2023internvl}
Chen, Z., Wu, J., Wang, W., Su, W., Chen, G., Xing, S., Zhong, M., Zhang, Q.,
  Zhu, X., Lu, L., et~al.: Internvl: Scaling up vision foundation models and
  aligning for generic visual-linguistic tasks. In: Proceedings of the IEEE/CVF
  conference on computer vision and pattern recognition. pp. 24185--24198
  (2024)

\bibitem{dai2023saferlhf}
Dai, J., Pan, X., Sun, R., Ji, J., Xu, X., Liu, M., Wang, Y., Yang, Y.: Safe
  rlhf: Safe reinforcement learning from human feedback (2023),
  \url{https://arxiv.org/abs/2310.12773}, accessed: 26 Jun 2026

\bibitem{feng2025video}
Feng, K., Gong, K., Li, B., Guo, Z., Wang, Y., Peng, T., Wu, J., Zhang, X.,
  Wang, B., Yue, X.: Video-r1: Reinforcing video reasoning in mllms (2025),
  \url{https://arxiv.org/abs/2503.21776}, accessed: 26 Jun 2026

\bibitem{fu2025video}
Fu, C., Dai, Y., Luo, Y., Li, L., Ren, S., Zhang, R., Wang, Z., Zhou, C., Shen,
  Y., Zhang, M., et~al.: Video-mme: The first-ever comprehensive evaluation
  benchmark of multi-modal llms in video analysis. In: Proceedings of the
  Computer Vision and Pattern Recognition Conference. pp. 24108--24118 (2025)

\bibitem{gemini2024}
{Gemini Team}: Gemini 1.5: Unlocking multimodal understanding across millions
  of tokens of context. arXiv preprint arXiv:2403.05530  (2024),
  \url{https://arxiv.org/abs/2403.05530}, accessed: 26 Jun 2026

\bibitem{google2025gemini}
{Google DeepMind}: Introducing gemini 2.0: Our new ai model for the agentic
  era.
  \url{https://blog.google/technology/google-deepmind/google-gemini-ai-update-december-2024/}
  (Dec 2024), accessed: 26 Jun 2026

\bibitem{gunjal2025rubrics}
Gunjal, A., Wang, A., Lau, E., Nath, V., He, Y., Liu, B., Hendryx, S.: Rubrics
  as rewards: Reinforcement learning beyond verifiable domains (2025),
  \url{https://arxiv.org/abs/2507.17746}, accessed: 26 Jun 2026

\bibitem{guo2025deepseek}
Guo, D., Yang, D., Zhang, H., Song, J., Wang, P., Zhu, Q., Xu, R., Zhang, R.,
  Ma, S., Bi, X., et~al.: {DeepSeek-R1} incentivizes reasoning in {LLMs}
  through reinforcement learning. Nature  \textbf{645}(8081),  633--638
  (September 2025). \doi{10.1038/s41586-025-09422-z},
  \url{https://doi.org/10.1038/s41586-025-09422-z}, accessed: 26 Jun 2026

\bibitem{he2025rifl}
He, Y., Li, W., Zhang, H., Li, S., Mandyam, K., Khosla, S., Xiong, Y., Wang,
  N., Peng, X., Li, B., Bi, S., Patil, S.G., Qi, Q., Feng, S., Katz-Samuels,
  J., Pang, R.Y., Gonugondla, S., Lang, H., Yu, Y., Qian, Y., Fazel-Zarandi,
  M., Yu, L., Benhalloum, A., Awadalla, H., Faruqui, M.: Advancedif:
  Rubric-based benchmarking and reinforcement learning for advancing llm
  instruction following (2025), \url{https://arxiv.org/abs/2511.10507},
  accessed: 26 Jun 2026

\bibitem{hu2022lora}
Hu, E.J., Shen, Y., Wallis, P., Allen-Zhu, Z., Li, Y., Wang, S., Wang, L.,
  Chen, W.: Lora: Low-rank adaptation of large language models (2021),
  \url{https://arxiv.org/abs/2106.09685}, accessed: 26 Jun 2026

\bibitem{jia2026openrs}
Jia, R., Yang, Y., Wu, Y., Gai, Y., Tao, S., Zhou, M., Lin, J., Jiang, X.,
  Jiang, G.: Open rubric system: Scaling reinforcement learning with pairwise
  adaptive rubric (2026), \url{https://arxiv.org/abs/2602.14069}, accessed: 26
  Jun 2026

\bibitem{jin2025omnireward}
Jin, Z., Yuan, H., Zhu, K., Li, J., Cao, P., Chen, Y., Liu, K., Zhao, J.:
  Omni-reward: Towards generalist omni-modal reward modeling with free-form
  preferences (2025), \url{https://arxiv.org/abs/2510.23451}, accessed: 26 Jun
  2026

\bibitem{kaufmann2023rlhf}
Kaufmann, T., Weng, P., Bengs, V., H{\"u}llermeier, E.: A survey of
  reinforcement learning from human feedback. arXiv preprint arXiv:2312.14925
  (2023), \url{https://arxiv.org/abs/2312.14925}, accessed: 26 Jun 2026

\bibitem{kirstain2023pickapic}
Kirstain, Y., Polyak, A., Singer, U., Matiana, S., Penna, J., Levy, O.:
  Pick-a-pic: An open dataset of user preferences for text-to-image generation.
  Advances in neural information processing systems  \textbf{36},  36652--36663
  (2023)

\bibitem{lee2023rlaif}
Lee, H., Phatale, S., Mansoor, H., Mesnard, T., Ferret, J., Lu, K., Bishop, C.,
  Hall, E., Carbune, V., Rastogi, A., Prakash, S.: Rlaif vs. rlhf: Scaling
  reinforcement learning from human feedback with ai feedback (2024),
  \url{https://arxiv.org/abs/2309.00267}, accessed: 26 Jun 2026

\bibitem{li2024llavaonevision}
Li, B., Zhang, Y., Guo, D., Zhang, R., Li, F., Zhang, H., Zhang, K., Zhang, P.,
  Li, Y., Liu, Z., Li, C.: Llava-onevision: Easy visual task transfer (2024),
  \url{https://arxiv.org/abs/2408.03326}, accessed: 26 Jun 2026

\bibitem{Li2025vlreward}
Li, L., Wei, Y., Xie, Z., Yang, X., Song, Y., Wang, P., An, C., Liu, T., Li,
  S., Lin, B.Y., Kong, L., Liu, Q.: Vl-rewardbench: A challenging benchmark for
  vision-language generative reward models. In: Proceedings of the IEEE/CVF
  Conference on Computer Vision and Pattern Recognition (CVPR). pp.
  24657--24668 (June 2025)

\bibitem{li2026rubrichub}
Li, S., Zhao, J., Wei, M., Ren, H., Zhou, Y., Yang, J., Liu, S., Zhang, K.,
  Chen, W.: Rubrichub: A comprehensive and highly discriminative rubric dataset
  via automated coarse-to-fine generation (2026),
  \url{https://arxiv.org/abs/2601.08430}, accessed: 26 Jun 2026

\bibitem{liu2024llavanext}
Liu, H., Li, C., Li, Y., Li, B., Zhang, Y., Shen, S., Lee, Y.J.: Llava-next:
  Improved reasoning, ocr, and world knowledge.
  \url{https://llava-vl.github.io/blog/2024-01-30-llava-next/} (2024),
  accessed: 26 Jun 2026

\bibitem{liu2023visualinstructiontuning}
Liu, H., Li, C., Wu, Q., Lee, Y.J.: Visual instruction tuning. Advances in
  neural information processing systems  \textbf{36},  34892--34916 (2023)

\bibitem{liu2026open}
Liu, T., Xu, R., Yu, T., Hong, I., Yang, C., Zhao, T., Wang, H.: Openrubrics:
  Towards scalable synthetic rubric generation for reward modeling and llm
  alignment (2026), \url{https://arxiv.org/abs/2510.07743}, accessed: 26 Jun
  2026

\bibitem{lu2025schemarl}
Lu, Y., Li, H., Cong, X., Zhang, Z., Wu, Y., Lin, Y., Liu, Z., Liu, F., Sun,
  M.: Learning to generate structured output with schema reinforcement learning
  (2025), \url{https://arxiv.org/abs/2502.18878}, accessed: 26 Jun 2026

\bibitem{openai2024gpt4o}
{OpenAI}: Hello gpt-4o: Introducing our real-time multimodal model.
  \url{https://openai.com/index/hello-gpt-4o/} (2024), accessed: 26 Jun 2026

\bibitem{ouyang2022training}
Ouyang, L., Wu, J., Jiang, X., Almeida, D., Wainwright, C., Mishkin, P., Zhang,
  C., Agarwal, S., Slama, K., Ray, A., et~al.: Training language models to
  follow instructions with human feedback. Advances in neural information
  processing systems  \textbf{35},  27730--27744 (2022)

\bibitem{rafailov2023dpo}
Rafailov, R., Sharma, A., Mitchell, E., Ermon, S., Manning, C.D., Finn, C.:
  Direct preference optimization: Your language model is secretly a reward
  model (2024), \url{https://arxiv.org/abs/2305.18290}, accessed: 26 Jun 2026

\bibitem{shao2024deepseek}
Shao, Z., Wang, P., Zhu, Q., Xu, R., Song, J., Bi, X., Zhang, H., Zhang, M.,
  Li, Y.K., Wu, Y., Guo, D.: Deepseekmath: Pushing the limits of mathematical
  reasoning in open language models (2024),
  \url{https://arxiv.org/abs/2402.03300}, accessed: 26 Jun 2026

\bibitem{srivastava2025crome}
Srivastava, P., Singh, H., Madhavan, R., Patil, G., Addepalli, S., Suggala, A.,
  Aravamudhan, R., Sharma, S., Laha, A., Raghuveer, A., Shanmugam, K., Precup,
  D.: Robust reward modeling via causal rubrics (2025),
  \url{https://arxiv.org/abs/2506.16507}, accessed: 26 Jun 2026

\bibitem{sun2024aligning}
Sun, Z., Shen, S., Cao, S., Liu, H., Li, C., Shen, Y., Gan, C., Gui, L.Y.,
  Wang, Y.X., Yang, Y., et~al.: Aligning large multimodal models with factually
  augmented rlhf (2024), \url{https://arxiv.org/abs/2309.14525}, accessed: 26
  Jun 2026

\bibitem{wang2026earlytom}
Wang, H., Jin, X., Lu, L., Li, C., Chen, J., Liu, Q., Wang, H.: Earlytom: Early
  token compression completes fast video understanding (2026),
  \url{https://arxiv.org/abs/2605.30010}, accessed: 26 Jun 2026

\bibitem{wang2025vrthinker}
Wang, Q., Liu, J., Liang, J., Jiang, Y., Zhang, Y., Chen, J., Zheng, Y., Wang,
  X., Wan, P., Yue, X., Liu, J., et~al.: Vr-thinker: Boosting video reward
  models through thinking-with-image reasoning. arXiv preprint arXiv:2510.10518
   (2025), \url{https://arxiv.org/abs/2510.10518}, accessed: 26 Jun 2026

\bibitem{wang2025unifiedrewardthink}
Wang, Y., Li, Z., Zang, Y., Wang, C., Lu, Q., Jin, C., Wang, J.: Unified
  multimodal chain-of-thought reward model through reinforcement fine-tuning.
  arXiv preprint arXiv:2505.03318  (2025),
  \url{https://arxiv.org/abs/2505.03318}, accessed: 26 Jun 2026

\bibitem{wang2024helpsteer2}
Wang, Z., Dong, Y., Delalleau, O., Zeng, J., Shen, G., Egert, D., Zhang, J.,
  Sreedhar, M.N., Kuchaiev, O.: Helpsteer 2: Open-source dataset for training
  top-performing reward models. Advances in Neural Information Processing
  Systems  \textbf{37},  1474--1501 (2024)

\bibitem{xie2026autorubric}
Xie, L., Huang, S., Zhang, Z., Zou, A., Zhai, Y., Ren, D., Zhang, K., Hu, H.,
  Liu, B., Chen, H., Liu, Z., Ding, B.: Auto-rubric: Learning from implicit
  weights to explicit rubrics for reward modeling (2026),
  \url{https://arxiv.org/abs/2510.17314}, accessed: 26 Jun 2026

\bibitem{xiong2025llava}
Xiong, T., Wang, X., Guo, D., Ye, Q., Fan, H., Gu, Q., Huang, H., Li, C.:
  Llava-critic: Learning to evaluate multimodal models. In: Proceedings of the
  Computer Vision and Pattern Recognition Conference. pp. 13618--13628 (2025)

\bibitem{xu2023imagereward}
Xu, J., Liu, X., Wu, Y., Tong, Y., Li, Q., Ding, M., Tang, J., Dong, Y.:
  Imagereward: Learning and evaluating human preferences for text-to-image
  generation. Advances in Neural Information Processing Systems  \textbf{36},
  15903--15935 (2023)

\bibitem{xu2025qwen25omni}
Xu, J., Guo, Z., He, J., Hu, H., He, T., Bai, S., Chen, K., Wang, J., Fan, Y.,
  Dang, K., Zhang, B., Wang, X., Chu, Y., Lin, J.: Qwen2.5-omni technical
  report (2025), \url{https://arxiv.org/abs/2503.20215}, accessed: 26 Jun 2026

\bibitem{xu2026positionbias}
Xu, Y., Hirasawa, T., Kozuno, T., Ushiku, Y.: Am i more pointwise or pairwise?
  revealing position bias in rubric-based llm-as-a-judge (2026),
  \url{https://arxiv.org/abs/2602.02219}, accessed: 26 Jun 2026

\bibitem{yang2022avqa}
Yang, P., Wang, X., Duan, X., Chen, H., Hou, R., Jin, C., Zhu, W.: Avqa: A
  dataset for audio-visual question answering on videos. In: Proceedings of the
  30th ACM international conference on multimedia. pp. 3480--3491 (2022)

\bibitem{yang2025r1}
Yang, Y., He, X., Pan, H., Jiang, X., Deng, Y., Yang, X., Lu, H., Yin, D., Rao,
  F., Zhu, M., et~al.: R1-onevision: Advancing generalized multimodal reasoning
  through cross-modal formalization. arXiv preprint arXiv:2503.10615  (2025)

\bibitem{yasunaga2025multimodal}
Yasunaga, M., Zettlemoyer, L., Ghazvininejad, M.: Multimodal rewardbench:
  Holistic evaluation of reward models for vision language models (2025),
  \url{https://arxiv.org/abs/2502.14191}, accessed: 26 Jun 2026

\bibitem{yue2024mmmu}
Yue, X., Ni, Y., Zhang, K., Zheng, T., Liu, R., Zhang, G., Stevens, S., Jiang,
  D., Ren, W., Sun, Y., et~al.: Mmmu: A massive multi-discipline multimodal
  understanding and reasoning benchmark for expert agi. In: Proceedings of the
  IEEE/CVF Conference on Computer Vision and Pattern Recognition. pp.
  9556--9567 (2024)

\bibitem{zhang2026chasingtail}
Zhang, J., Wang, Z., Gui, L., Sathyendra, S.M., Jeong, J., Veitch, V., Wang,
  W., He, Y., Liu, B., Jin, L.: Chasing the tail: Effective rubric-based reward
  modeling for large language model post-training (2026),
  \url{https://arxiv.org/abs/2509.21500}, accessed: 26 Jun 2026

\bibitem{zhang2024direct}
Zhang, R., Gui, L., Sun, Z., Feng, Y., Xu, K., Zhang, Y., Fu, D., Li, C.,
  Hauptmann, A., Bisk, Y., Yang, Y.: Direct preference optimization of video
  large multimodal models from language model reward. arXiv preprint
  \textbf{2404.01258} (2024), \url{https://arxiv.org/abs/2404.01258}, accessed:
  26 Jun 2026

\bibitem{zhao2025r1reward}
Zhang, Y.F., Lu, X., Hu, X., Fu, C., Wen, B., Zhang, T., Liu, C., et~al.:
  {R1-Reward: Training Multimodal Reward Model through Stable Reinforcement
  Learning}. arXiv preprint  \textbf{2505.02835} (2025),
  \url{https://arxiv.org/abs/2505.02835}, accessed: 26 Jun 2026

\bibitem{zhang2025baserewardstrongbaselinemultimodal}
Zhang, Y.F., Yang, H., Zhang, H., Shi, Y., Chen, Z., Tian, H., Fu, C., Wang,
  H., Wu, K., Cui, B., Wang, X., Pan, J., Wang, H., Zhang, Z., Wang, L.:
  Basereward: A strong baseline for multimodal reward model (2025),
  \url{https://arxiv.org/abs/2509.16127}, accessed: 26 Jun 2026

\bibitem{zhou2026ruscarl}
Zhou, Y., Li, S., Liu, S., Fang, W., Zhang, K., Zhao, J., Yang, J., Zhou, Y.,
  Lv, J., Zheng, T., Lu, H., Chen, W., Xie, Y., Song, M.: Breaking the
  exploration bottleneck: Rubric-scaffolded reinforcement learning for general
  llm reasoning (2026), \url{https://arxiv.org/abs/2508.16949}, accessed: 26
  Jun 2026

\bibitem{zhu2025internvl3}
Zhu, J., Wang, W., Chen, Z., et~al.: Internvl3: Exploring advanced training and
  test-time recipes for open-source multimodal models (2025),
  \url{https://arxiv.org/abs/2504.10479}, accessed: 26 Jun 2026

\end{thebibliography}
\end{document}